%% file: main.tex
\documentclass{article}

\usepackage[preprint]{corl_2023} 
\include{include}
\include{macro}

\usepackage{multirow}

\title{Reinforcement Learning with Human Feedback for Realistic Traffic Simulation}

\author{
  Yulong Cao \\
  NVIDIA\\
  \And
  Boris Ivanovic \\
  NVIDIA\\
  \And
  Chaowei Xiao \\
 NVIDIA, UW-Madison\\
  \And
  Marco Pavone \\
  NVIDIA, Stanford\\
}

\begin{document}
\maketitle


\begin{abstract}
    In light of the challenges and costs of real-world testing, autonomous vehicle developers often rely on testing in simulation for the creation of reliable systems. A key element of effective simulation is the incorporation of realistic traffic models that align with human knowledge, an aspect that has proven challenging due to the need to balance realism and diversity. This works aims to address this by developing a framework that employs reinforcement learning with human preference (RLHF) to enhance the realism of existing traffic models. This study also identifies two main challenges: capturing the nuances of human preferences on realism and the unification of diverse traffic simulation models. To tackle these issues, we propose using human feedback for alignment and employ RLHF due to its sample efficiency. We also introduce the first dataset for realism alignment in traffic modeling to support such research.
    Our framework, named TrafficRLHF, demonstrates its proficiency in generating realistic traffic scenarios that are well-aligned with human preferences, as corroborated by comprehensive evaluations on the nuScenes dataset. 
\end{abstract}



\input{body/1_intro}
\input{body/2_related}

\input{body/3_method}
\input{body/4_experiment}
\input{body/5_conclusion}




\bibliography{references}
\clearpage
\input{body/appendix}

\end{document}

%% file: include.tex
\usepackage{graphicx}
\usepackage{wrapfig}
\usepackage{amsmath}
\usepackage{amsfonts}
\usepackage{cleveref}
\usepackage{enumitem}
\usepackage{comment}
\usepackage{subfig}
\usepackage{xspace}
\usepackage{xcolor}
\usepackage{tabularx}
\usepackage{bm}
\usepackage{algorithm}
\usepackage{algorithmic}
\usepackage{booktabs}
\usepackage{MnSymbol}

%% file: macro.tex
\newcommand{\method}{TrafficRLHF\xspace}

\definecolor{tblue}{HTML}{1F77B4}
\definecolor{tred}{HTML}{FF6961}
\definecolor{tgreen}{HTML}{429E9D}
\definecolor{thighlight}{HTML}{000000}
\newcolumntype{P}{>{\raggedleft\arraybackslash}X}

\definecolor{cred}{HTML}{D62728}
\definecolor{cblue}{HTML}{1F77B4}
\definecolor{cgreen}{HTML}{79AB76}
\definecolor{cgrey}{rgb}{0.6,0.6,0.6}

\definecolor{highlight}{rgb}{0,0,0}

\newcommand{\bc}{\mathbf{c}}

%% file: body/1_intro.tex
\section{Introduction}
Due to the significant expenses and risks associated with conducting large-scale real-world tests~\cite{nummilesneeded}, autonomous vehicle (AV) developers rely heavily on comprehensive testing in simulation to ensure the development of reliable systems~\cite{waymoreport}.
To maximize the effectiveness of simulators, it is crucial for them to offer \textit{realism} that is aligned with human knowledge.
Accordingly, \textit{realistic} traffic models are essential to ensure that insights gained from simulation testing apply seamlessly to real-world scenarios~\cite{Suo_2021_CVPR,bits2022}. However, developing traffic models that balance realism and diversity remains an ongoing challenge. To tackle this challenge, we leverage recent advancements in alignment research for traffic modeling; moreover, such alignment is model-agnostic and can improve various existing traffic models. Consequently, our primary research objective is to develop a framework based on reinforcement learning with human feedback (RLHF) to widely improve the realism of existing traffic models.

In order to align human preferences for generating realistic traffic simulations, two significant challenges need to be addressed: (1) the limited expressiveness of existing methods in capturing human preferences for realism in traffic simulations, and (2) unifying diverse traffic simulation models.
The first challenge relates to traditional approaches relying on predefined rules or statistical models that fail to encompass the wide range of human preferences and subjective perceptions of realistic traffic scenarios. 
The second challenge revolves around the need to unify diverse traffic simulation models. The field of traffic simulation encompasses a wide range of models, each with their own strengths and limitations, underlying assumptions, data requirements, and simulation algorithms, making it challenging to incorporate human preferences seamlessly. 

\begin{figure}[t]
    \centering
    \includegraphics[width=0.95\textwidth]{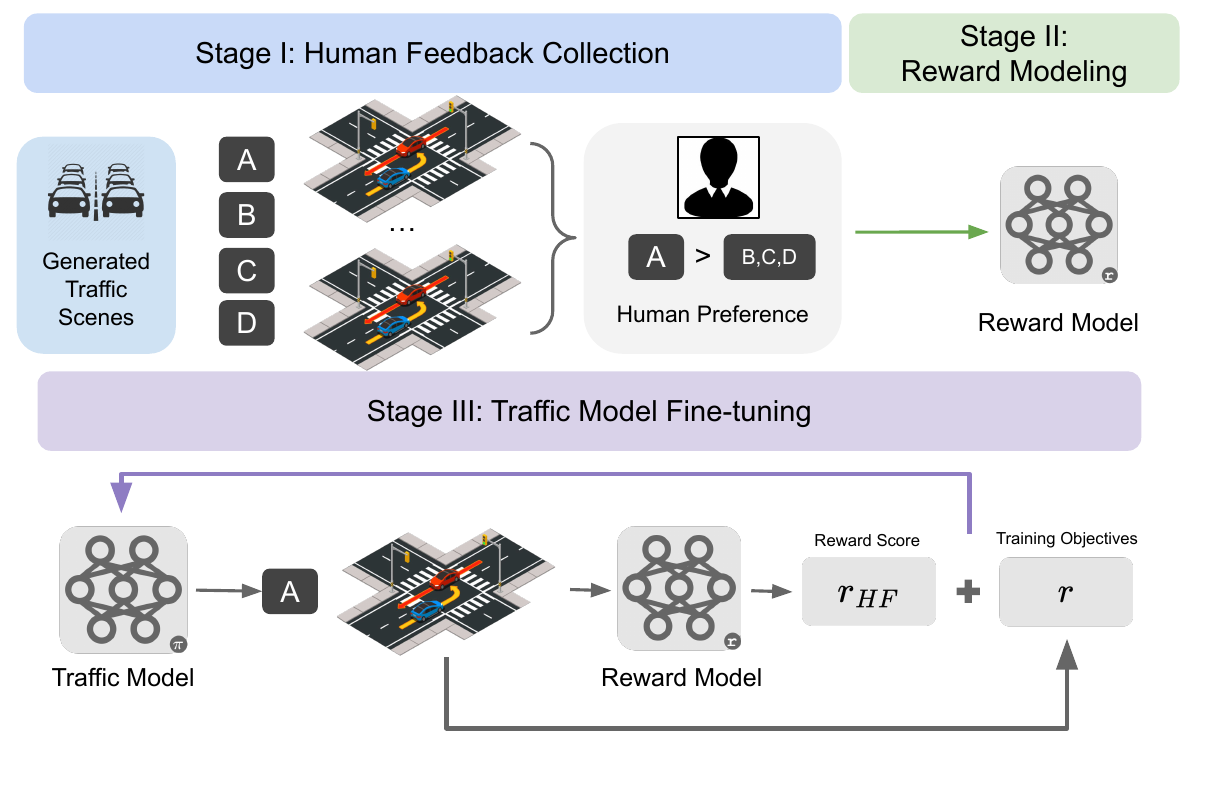}

    \vspace{-0.5cm}
    
    \caption{An overview of our 3-stage \method approach. Stage I entails the collection of human feedback on traffic scenarios to train a reward model. In Stage II, this reward model is utilized to rank the realism of traffic scenarios. Lastly, Stage III involves the fine-tuning of the traffic models based on the reward model, thereby improving the realism of the generated traffic scenarios.}
    \label{fig:method_overview}
\end{figure}

To address these challenges, we introduce \method, a three-stage framework utilizing human feedback to enhance traffic models for realistic scenario generation. The framework tackles key challenges in the field of realistic traffic simulation, leveraging recent advancements in reinforcement learning with human feedback (RLHF) and a state-of-the-art autoregressive backbone model.

To tackle the first challenge—capturing human preferences on realism—we propose using human feedback as a tool for alignment. RLHF has recently demonstrated its effectiveness in conjunction with large language models. Its impressive sample efficiency makes it a prime candidate for realism alignment in traffic models. Moreover, through the labeling of a modest quantity of data with human preferences, one can establish a reward model suitable for refining a wide range of traffic models.
Addressing the second challenge—the need to unify diverse traffic simulation models—requires a method that can be applied across diverse model architectures. This is achieved using our RLHF-based method, which only requires a universal interface for reward model input. To fulfill this requirement, we employ a state-of-the-art autoregressive backbone model, CTG~\cite{zhong2022guided}, with a preferred roll-out length to optimize the simulation of the near future.
Experiments on the real-world nuScenes autonomous driving dataset~\cite{nuscenes} confirm \method's ability to generate realistic trajectories that align with human preferences. As one result, \method is able to reduce unrealistic collisions or off-road driving by up to 80\%. 

\textbf{Contributions.} The main contributions of our work are threefold: \textbf{(1)} We introduce the first dataset designed for realism alignment in traffic modeling, \textbf{(2)} we formulate a reward model that quantifies realism in alignment with human preferences, and \textbf{(3)} we propose a model-agnostic framework, leveraging RLHF, that effectively enhances the realism of a broad range of existing traffic models.

%% file: body/2_related.tex
\section{Related Work}
\noindent\textbf{Traffic simulation.}
Traffic simulation techniques can be broadly categorized into rule-based and learning-based approaches. Rule-based methods, employing analytical models such as cellular automata and intelligent driver models~\cite{microtrafficmodelbenchmark03}, tend to set fixed routes for vehicles and separate longitudinal from lateral agent movements. This inflexibility restricts their capacity to mimic the dynamic variety inherent in real-world driving behaviors.

On the other hand, learning-based approaches utilize deep generative models trained on trajectory datasets to mimic real-world driving behaviors~\cite{chai2019multipath,Chen_2022_CVPR,salzmann2020trajectron++, Suo_2021_CVPR, bits2022}. However, they often lack the flexibility for users to dictate customized attributes of generated traffic behaviors during inference, thereby falling short in capturing the full spectrum of realistic traffic scenarios.

Additionally, certain researches emphasize on generating adversarial or safety-critical scenarios, crafting trajectories that provoke autonomous vehicle misbehavior~\cite{wang2021advsim, advdo, Abeysirigoonawardena2019Generating, chenbaiming2020}. Models like STRIVE~\cite{rempe2022strive} and CTG~\cite{zhong2022guided}, and TrafficGen~\cite{feng2022trafficgen} have made strides in this direction. But, these models often struggle to generate traffic scenarios that align with human perceptions of realism.

Despite the advancements in traffic simulation methodologies, a significant gap persists in the field: the ability to produce traffic scenarios that closely align with human perceptions of realism, an essential component for effective AV testing. Therefore, integrating human preferences seamlessly into traffic simulation models remains an unfulfilled challenge and a key focus of our research.

\noindent\textbf{Reinforcement Learning from Human Feedback.}
The investigation into Reinforcement Learning from Human Feedback (RLHF) spans at least a decade, with significant contributions from numerous researchers \cite{akrour2011preference,akrour2012april,griffith2013policy,christiano2017deep,jaques2019way}. RLHF is a component of a more comprehensive paradigm known as human-in-the-loop learning process \cite{wu2022survey}. In scenarios where reward engineering for RL proves challenging or costly, human feedback data emerges as a crucial asset.

Recent advances in the field have seen the use of human feedback for fine-tuning LLMs \cite{ziegler2019fine,jaques2019way,stiennon2020learning}. A notable example is InstructGPT \cite{ouyang2022training}, which leverages both human demonstrations and preferences to achieve significant improvements over the GPT-3 baseline in terms of human preferences from annotators. This demonstrates the scalability and potential of RLHF as a method for tuning large models using human feedback.

Our proposed RLHF approach offers unique advantages for generating realistic traffic scenarios with human feedback, particularly in terms of realism that lack of explicit formulations. This approach minimizes data collection and feedback time, avoiding the need for costly labeled data. The preference for videos requires minimal human effort, thus enabling the production of a large volume of labeled data.

%% file: body/3_method.tex
\section{TrafficRLHF}
\label{sec:method}

In this section, we formally specify the traffic simulation problem and delineate the three core components of \method. As shown in Figure~\ref{fig:method_overview}, these components include data collection, involving the accumulation of human preferences for realism using a blend of genuine traffic scenes and those generated from traffic models; reward model training, which entails training a model capable of evaluating the realism of a traffic scenario based on human preferences; and fine-tuning, wherein we harmonize the model with the reward model trained in the previous stage. Ultimately, this methodology enables a wide array of traffic models to generate more authentic traffic scenarios.

\subsection{Traffic Simulation Formulation}
\label{sec:method-formulation}
We formulate the problem of traffic simulation similar to~\cite{zhong2022guided}. For a scenario with $M$ vehicles, their state (comprised of $x, y$ position, speed, and yaw) at timestep $t$ is represented as $s_{t}=[s_{t}^1~...s_{t}^M]$ where $s_{t}^i = (x_{t}^i,y_{t}^i, v_{t}^i, \theta_{t}^i)$, and their action (\textit{i.e.}, control) as $a_{t}=[a_{t}^1...~a_{t}^M]$, where $a_{t}^i=(\dot{v}{t},\dot{\theta}{t})$ (acceleration and yaw rate). We designate $\bc=(I,s_{t-T_\text{hist}:t+1})$ as decision-relevant context, comprising local semantic maps for all agents $I={I^1,...,I^M}$, as well as their current and $T_{hist}$ preceding states $s_{t-T_\text{hist}:t+1}={s_{t-T_\text{hist}}, \dots, s_{t}}$.
The state $s_{t+1}$ at time $t+1$ is derived from a transition function $f$ that computes $s_{t+1}=f(s_{t}, a_{t})$ given the previous state $s_{t}$ and control $a_{t}$ following unicycle dynamics.

\subsection{Human Feedback Collection}
\label{sec:method-human_feedback}
Our proposed framework \method aligns human preferences with generated traffic scenarios under identical initial conditions. 
It requires only a moderate amount of video data collection and minimal human feedback. Since all trajectories are collected in simulation using existing traffic models, a considerable amount of data can be generated and enhanced iteratively. Furthermore, any human with driving experience can serve as a labeler after a brief familiarization with the user interface, eliminating the need for specialized expertise.

To garner human feedback, pairs of traffic scenarios $(S_1, S_2)$ are generated from the same context $\bc$. Each scenario $S_i = [s_{t-T_\text{hist}:t+T_\text{fut}}]_i, i=1,2$ represents stacked state sequences of length $T\text{fut}$. Human annotators are shown videos of these paired traffic scenarios, with maps and vehicles appropriately labeled. The annotators are then tasked with specifying which scenario appears more realistic.

As current traffic models often yield less realistic data, we increase the total amount of generated traffic scenarios to boost the likelihood of generating at least one realistic scenario. Moreover, since the generated scenarios are usually covering multi-modal future and not always comparable, we select the most realistic example instead of a pairwise comparison. Concretely, we generate $N$ traffic scenarios $S_1,\cdots,S_N$ and request human annotators to select the most realistic one from the set. This process results in $N-1$ scenario pairs with preference relationships $(S_i\succ S_j)$, where $S_i$ is the scenario chosen by the human annotator and $S_j$ is any other scenario.
In cases where the traffic model fails to generate realistic scenarios, annotators are provided the option to declare that none of the presented scenarios appear realistic. If this option is selected, we will use the ground truth scenario $S_\text{gt}$ as the preferred scenario, denoted as $(S_\text{gt} \succ S_j)$.


\subsection{Reward Modeling}
\label{sec:method-reward_model}
The training methodology for the reward model $r_{\text{HF}}(\cdot;\phi)$ is inspired by Abramson et al.~\cite{abramson2022improving}. Two traffic scenarios, generated under the same initial conditions, are compared and the reward model is trained with the following loss:
\begin{equation}
    l_\text{RM} = \mathbb{E}_{S_1, S_2 \sim \mathcal{D}}[-\log\sigma(r_\text{HF}(S_1;\phi) - r_\text{HF}(S_2;\phi))],S_1 \succ S_2.
\label{eq:rm_loss}
\end{equation}
Here, $r_\text{HF}(S_i;\phi)$ represents the reward model. The training dataset $\mathcal{D}$ is compiled using human preferences over traffic scenarios, generated with the existing traffic models (without any improvements). And $S_1, S_2\sim \mathcal{D} $ is a pair of scenarios sampled from the human labeled dataset. Note that $r_\text{HF}$ can be either an iterative model or an encoder-decoder model taking a fixed-length sequence. To maximize \method's ability to adapt a variety of traffic models during the final fine-tuning stage, we choose an iterative model that can take as input an arbitrary sequence length. To calculate the reward, we compute the average of the reward over a sequence $S_i$.

\subsection{Traffic Model Fine-tuning}
\label{sec:method-finetune}
Similar to the fine-tuning method proposed by Abramson et al.~\cite{abramson2022improving}, our objective is:
\[\Tilde{r} = r + \alpha \cdot r_\text{HF}(S),\]
where $r$ is the original loss for training the provided traffic model and $\alpha$ is a scaling term. This mixed objective is used to avoid overfitting the reward score and degrading task performance. In RLHF for training large language models (LLMs) (e.g., ChatGPT \cite{vemprala2023chatgpt}), usually the majority of the model parameters are frozen for the computational benefits. However, existing traffic models are usually much smaller than LLMs. In our experiments, we will show that fine-tuning the entire motion prediction model can lead to higher performance improvements.

%% file: body/4_experiment.tex
\section{Experiments}
\label{sec:experiments}
We conduct experiments to validate that (1) \method can generate more realistic traffic behaviors, and (2) the reward model is generalizable and can be reused for a diverse range of traffic models.

\subsection{Experimental Setup}
\label{sec:setup}
\noindent\textbf{Datasets.} 
The nuScenes dataset \cite{nuscenes} is a vast compilation of real-world driving data, totaling 5.5 hours of meticulously logged vehicle trajectories from two unique urban settings. It captures a myriad of driving scenarios, including instances of high traffic density. In our study, we use mainly the training split of the nuScenes dataset for data collection, reward model training and traffic model fine-tuning. Evaluation of the traffic models' performance is carried out on a randomly sampled subset of 100 scenes from the validation split.

\noindent\textbf{Metrics.}
    Following~\cite{zhong2022guided, bits2022}, we evaluate stability (\textit{i.e.}, avoiding collisions and off-road driving), realism of generated trajectories by each model. We evaluate \textit{stability} by reporting the failure rate (\textbf{fail}), measured as the average percentage of agents encountering a critical failure (\textit{i.e.}, a collision or road departure) in a scene. To assess \textit{realism}, we compare data statistics between generated traffic simulations and ground truth trajectories from the dataset by calculating the Wasserstein distance between their normalized histograms of driving profiles. We measure \textit{realism} using realism deviation (\textbf{real}) which is the average of realism values for three properties: longitudinal acceleration magnitude, lateral acceleration magnitude, and jerk. Also, since the reward model naturally captured the human preference and we measure the human preference score using the reward cost score (\textbf{reward cost}).

\noindent\textbf{Traffic Model Benchmarks.} 
Recent advancements in the field of traffic generation include three particularly noteworthy models. Firstly, the Conditional Traffic Generation model (\textbf{CTG}) is a model predicated upon the principles of conditional diffusion~\cite{zhong2022guided}. Secondly, the Bi-Level Imitation Learning System (\textbf{BITS}) is a model that utilises a two-tiered imitation learning strategy to approach traffic prediction tasks~\cite{bits2022}. Lastly, \textbf{TrafficGen} is an autoregressive generative model that leverages an encoder-decoder architecture, enabling the diverse initialization of traffic scenarios~\cite{feng2022trafficgen}. We also need the initialized model parameters for the fine-tuning (\Cref{sec:method-finetune}). Given that both CTG and BITS were initially trained on the nuScenes dataset~\cite{nuscenes}, we elected to use their original parameters. As for TrafficGen, we managed to reproduce its performance on the same dataset using the original implementation provided by the authors.

\subsection{Experimental Details}
\noindent\textbf{Traffic Generation and Data Collection.}
In the generation of traffic scenarios, we utilize the state-of-the-art Conditional Traffic Generation (CTG) model. CTG model is built with a encoder for traffic context and a diffuser model for traffic generation. As diffuser presents the concept of guidance, which allows for the sampling of trajectories from an unconditional diffusion model to achieve a specific pre-set objective, CTG use such guidance for controllable generation. For the data collection, we use CTG to generate traffic scenarios, operating under the guidance of a ``no collision'' principle, a feature that is evident in their outcomes. From the nuScenes dataset, we selected a subset of 500 scenes from the training split~\cite{nuscenes}. Each scene was processed through the CTG model to generate five unique scenarios. To ascertain human preferences, we presented these five scenarios to labelers, tasking them with identifying the most realistic scenario or indicating if all scenarios lacked realism. Consequently, we constructed our human preference dataset by pairing either the most realistic or the ground truth sample with another sample. This curated data was then employed in training the reward model, with the loss calculated according to Equation ~\ref{eq:rm_loss}.

\noindent\textbf{Reward Model and Training.}
In our experiments, the Reward Model (RM) is parameterized using the CTG encoder and fully-connected neural networks for the output layer. This selection of inputs for the RM could be seen as a universal module applicable to any traffic scenario. Given the status of CTG as a state-of-the-art model, we assume it offers superior expressiveness across diverse traffic scenes. Furthermore, the simplicity of the CTG encoder facilitates a straightforward training schema, thereby reducing overhead when compared to other models that incorporate more complex components, such as RL policy networks~\cite{bits2022, Chen_2022_CVPR}. Nevertheless, it is essential to note that the choice of the reward model is not restricted to our current selection and has the potential for expansion and improvement in the future.

\begin{figure}[h]
    \centering
    \includegraphics[width=0.7\textwidth]{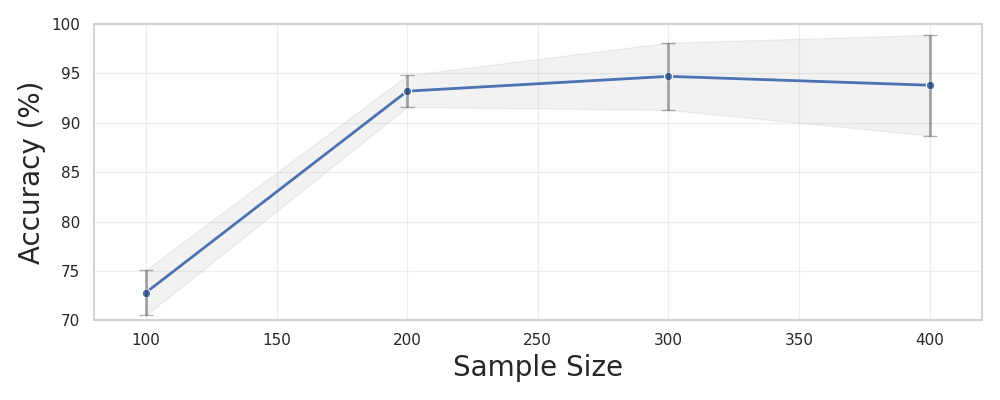}
\caption{Evaluating the accumulative rewards over trajectories with the RM with the human preferences.}
\label{fig:rm_val}
\end{figure}

\subsection{Reward Model Validation}
\label{sec:experiment-reward_model}
Before proceeding to fine-tune the traffic model, we first assess the Reward Model (RM) using the human preference data we collected. This evaluation involves comparing the reward scores with human preferences. For the evaluation, we partition the collected human preferences into a training subset and a validation subset, comprising 400 and 100 units respectively. As demonstrated in Figure~\ref{fig:rm_val}, the accuracy of the RM swiftly converges with an increase in sample size. However, the variance of the accuracy tends to broaden with further increments. This increasing variance potentially indicates overfitting during the training process~\cite{gao2022scaling}, since NuScenes dataset is relatively small in size ( 850 scenes). To optimize performance, we opt to utilize the RM trained with 200 samples in the fine-tuning process.

\begin{figure}[ht]
\centering
\includegraphics[width=0.24\textwidth,trim={4cm 4cm 1cm 1cm},clip]{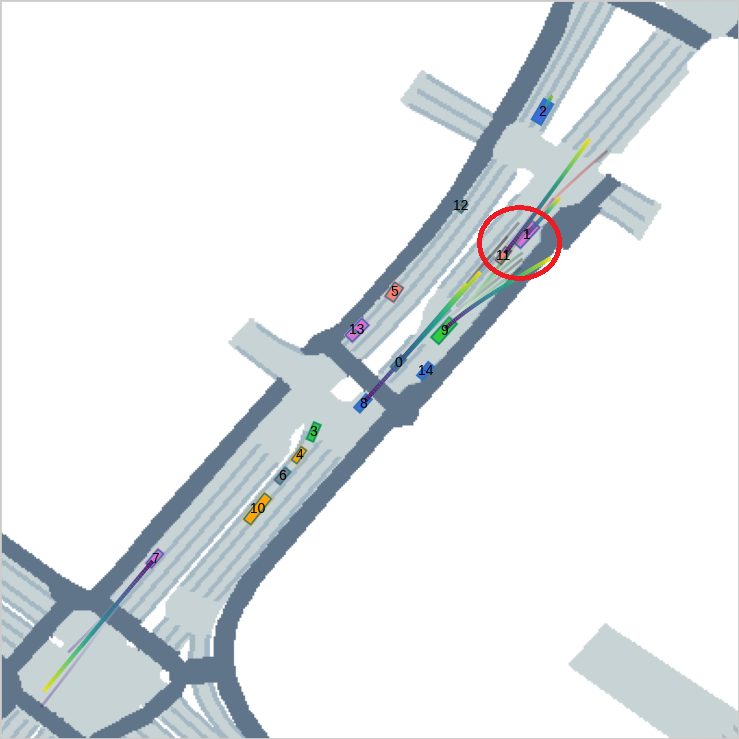} \hfill
\includegraphics[width=0.24\textwidth,trim={4cm 4cm 1cm 1cm},clip]{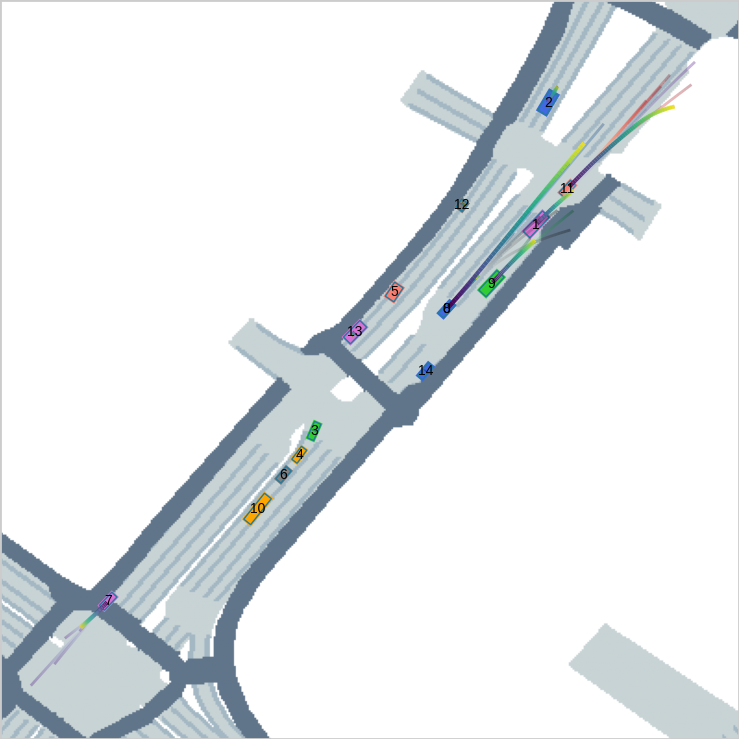} \hfill
\includegraphics[width=0.24\textwidth,trim={4cm 4cm 2cm 2cm},clip]{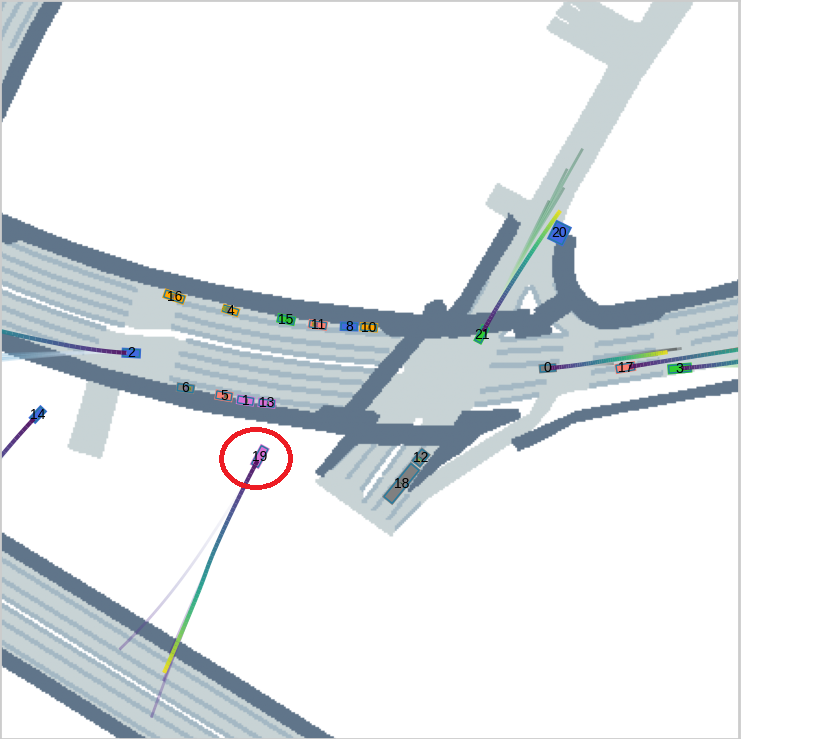} \hfill
\includegraphics[width=0.24\textwidth,trim={4cm 4cm 2cm 2cm},clip]{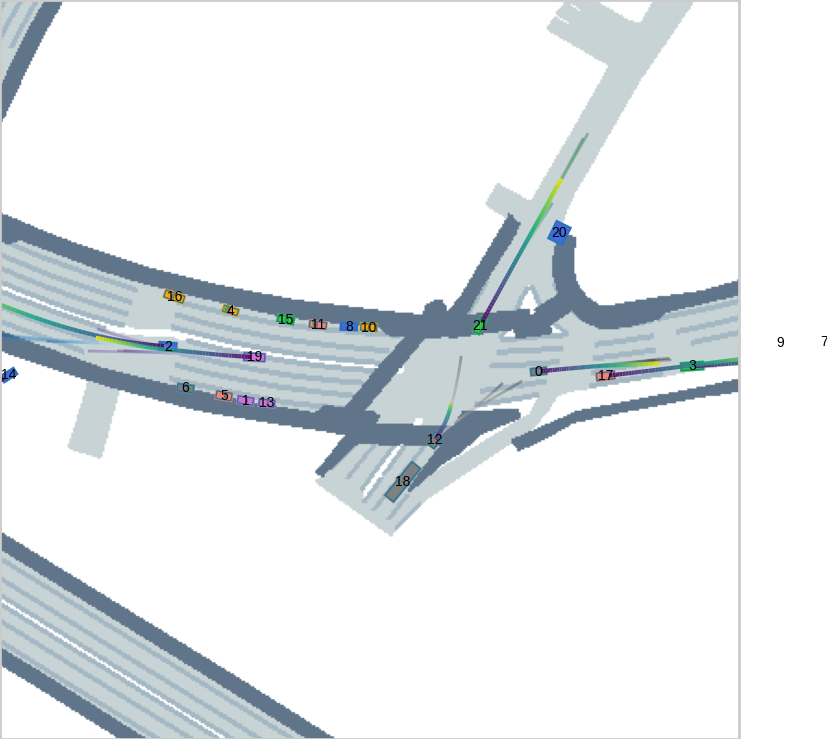} \hfill
\subfloat[][CTG collision \label{fig:ctg_example_collision}]{\includegraphics[width=0.24\textwidth,trim={2cm 2cm 2cm 2cm},clip]{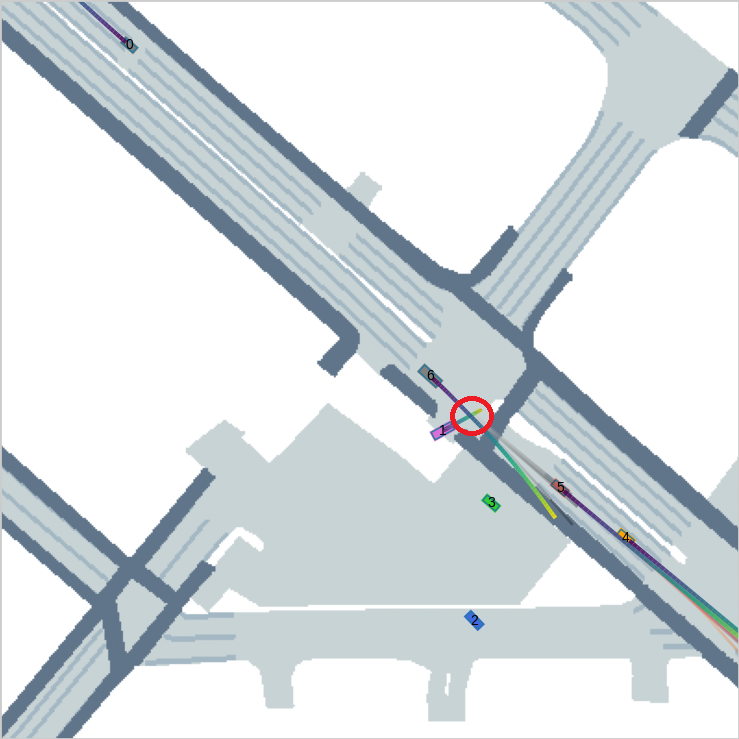}} \hfill
\subfloat[][CTG+RM no collision \label{fig:ctg_rlhf_example_collision}]{\includegraphics[width=0.24\textwidth,trim={2cm 2cm 2cm 2cm},clip]{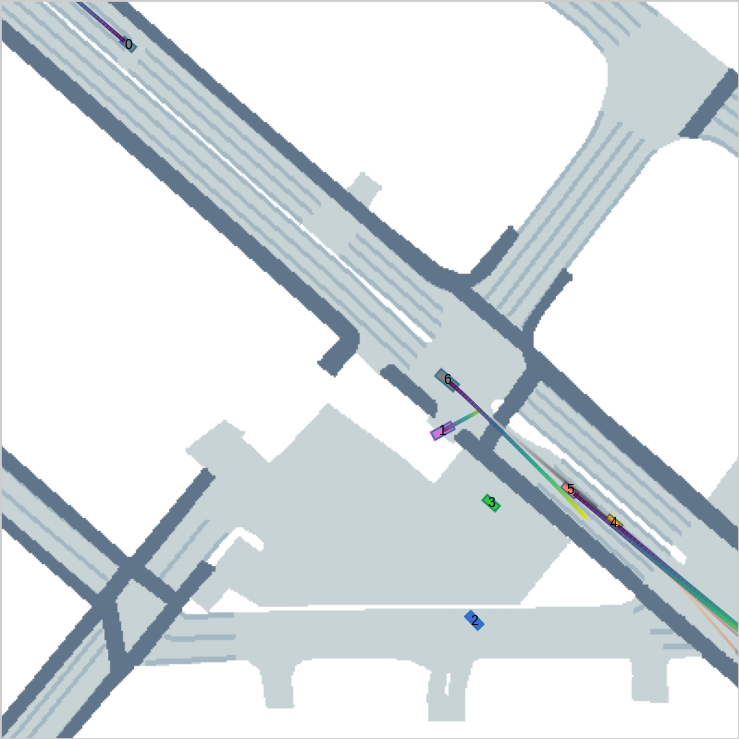}} \hfill
\subfloat[][CTG off-road \label{fig:ctg_example_oor}]{\includegraphics[width=0.24\textwidth,trim={3cm 3cm 1cm 1cm},clip]{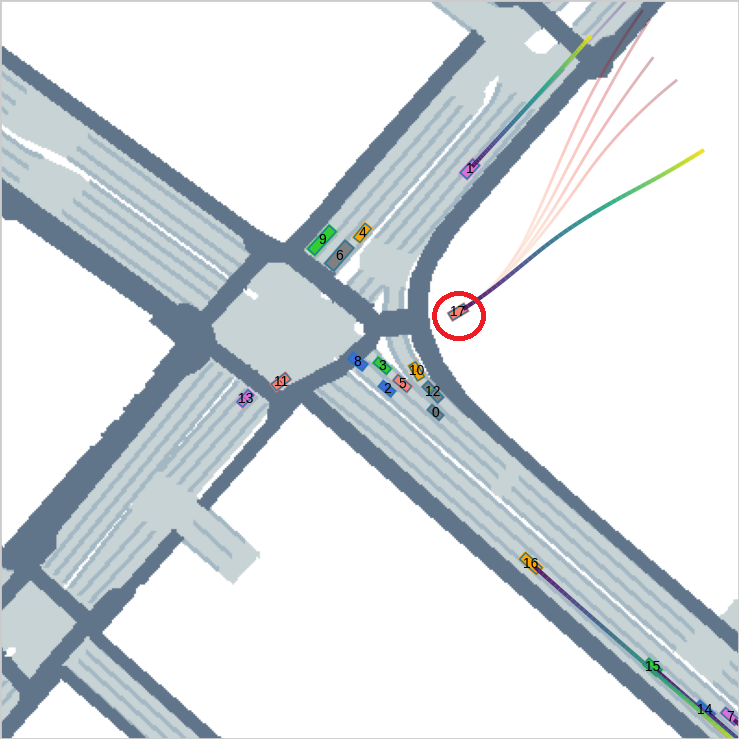}} \hfill
\subfloat[][CTG+RM stay in lane \label{fig:ctg_rlhf_example_oor}]{\includegraphics[width=0.24\textwidth,trim={3cm 3cm 1cm 1cm},clip]{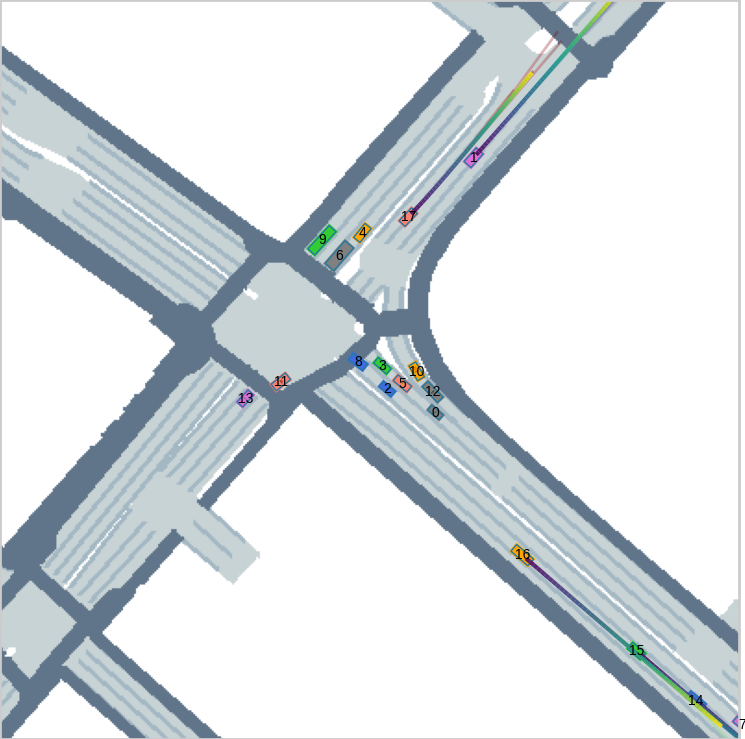}} \hfill
\caption{Qualitative results of CTG fine-tuned with RM avoids unrealistic collisions and off-road conditions. For better visibility, we marked the unrealistic behaviors generated in {\color{red}{red} circles}.}
\label{fig:ctg_example}
\end{figure}

\begin{table}[ht]
\centering
\caption{Quantitative results on CTG. The best results are highlighted.}
\begin{tabular}{c|cc|cc|cc|cc|cc}
\toprule 
       & \multicolumn{2}{c|}{\textbf{Speed limit}} & \multicolumn{2}{c|}{\textbf{Target speed}} & \multicolumn{2}{c|}{\textbf{No collision}} & \multicolumn{2}{c|}{\textbf{No off-road}} & \multicolumn{2}{c}{\textbf{Goal waypoint}} \\
       & real$\downarrow$                & fail$\downarrow$               & real$\downarrow$                & fail$\downarrow$                & real$\downarrow$                & fail$\downarrow$                & real$\downarrow$                & fail$\downarrow$               & real$\downarrow$                 & fail$\downarrow$                \\
       \midrule 
CTG    & \textbf{0.359}      & 0.165     & 0.855               & 0.179               & 0.569               & 0.271               & 0.501               & 0.455              & \textbf{0.564}       & 0.387               \\
CTG+RM & 0.361               & \textbf{0.152}              & \textbf{0.732}      & \textbf{0.156}      & \textbf{0.376}      & \textbf{0.054}      & \textbf{0.392}      & \textbf{0.214}     & 0.734                & \textbf{0.341}    \\
\bottomrule 
\end{tabular}
\label{tab:ctg_rm_results}
\end{table}

\subsection{Evaluation of Traffic Model}
\label{sec:traffic_model_eval}
\noindent\textbf{CTG+RM.}
Having evaluated the learned RM, we apply it to fine-tune the traffic model, as outlined in \Cref{sec:method-finetune}. As depicted in Figure~\ref{fig:ctg_example}, fine-tuning with the RM significantly enhances realism by decreasing the failure rates associated with collisions and driving off the road. For quantitative results, given that CTG offers a robust feature for controllable traffic scenario generation via guidance, we evaluate the model with varying guidance, following the original authors' setup. As demonstrated in Table~\ref{tab:ctg_rm_results}, fine-tuning enables the generated traffic scenarios to decrease failures (\textit{i.e.}, collisions and driving off the road). We observe that the realism metrics deteriorate when guidance is related to the goal positions/speeds, possibly because the guidance during the diffusion process in CTG overpowers the loss provided by the RM penalty losses.

\begin{figure}[t]
\centering
\subfloat[][\small{BITS collision} \label{fig:ctg_example_collision}]{\includegraphics[width=0.24\textwidth,trim={3cm 3cm 3cm 3cm},clip]{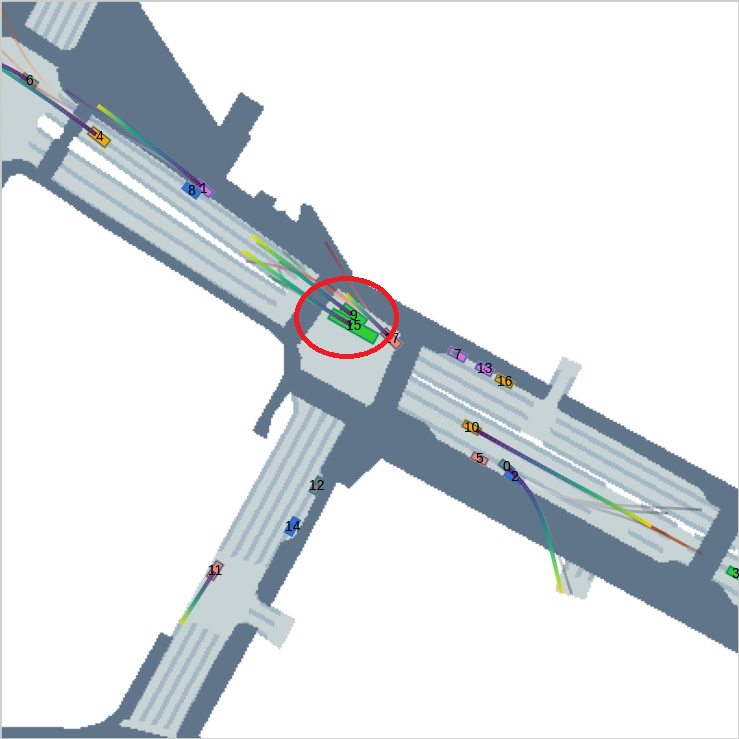}} \hfill
\subfloat[][\small{BITS+RM no collision} \label{fig:ctg_rlhf_example_collision}]{\includegraphics[width=0.24\textwidth,trim={3cm 3cm 3cm 3cm},clip]{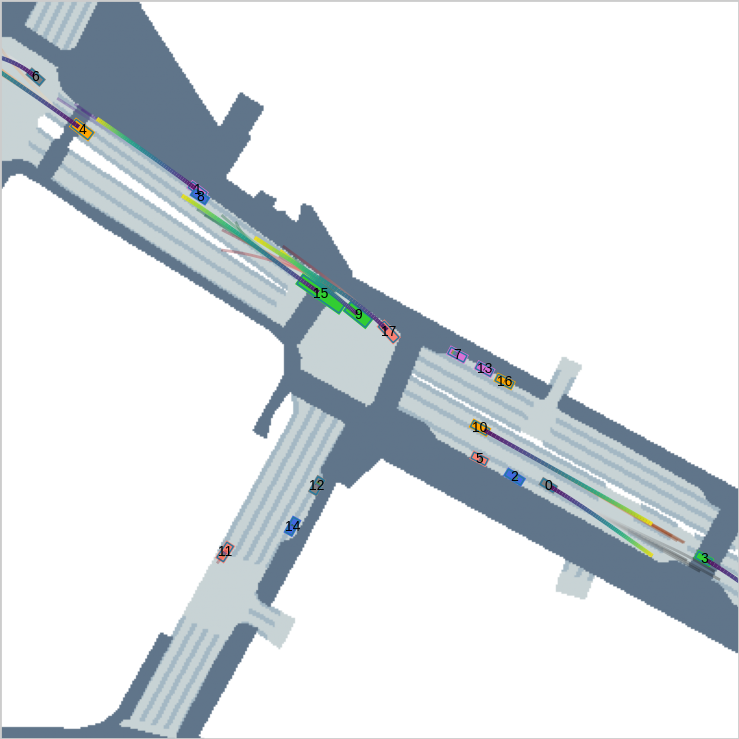}} \hfill
\subfloat[][\small{TrafficGen off-road} \label{fig:ctg_example_oor}]{\includegraphics[width=0.24\textwidth,trim={3cm 3cm 3cm 3cm},clip]{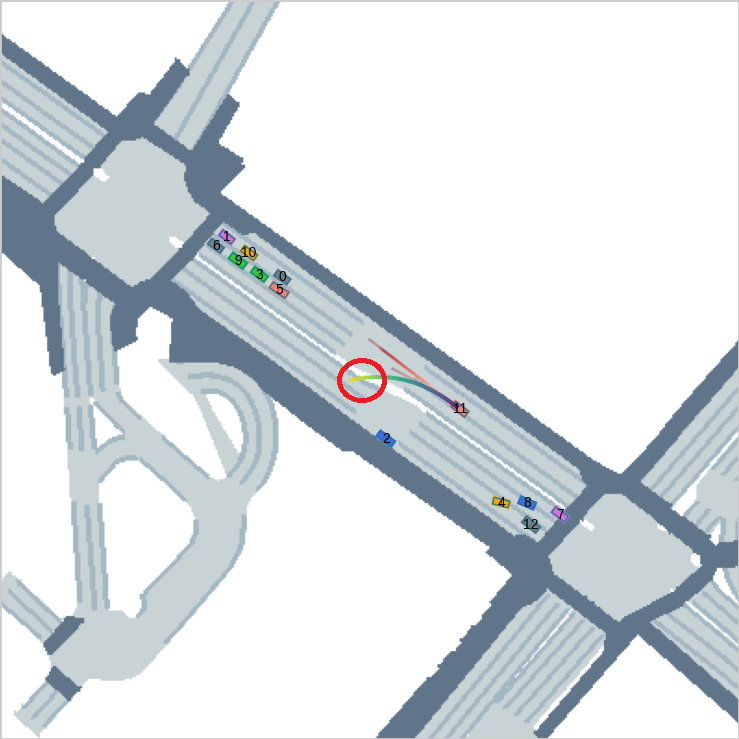}} \hfill
\subfloat[][\small{TrafficGen+RM stay in lane} \label{fig:ctg_rlhf_example_oor}]{\includegraphics[width=0.24\textwidth,trim={3cm 3cm 3cm 3cm},clip]{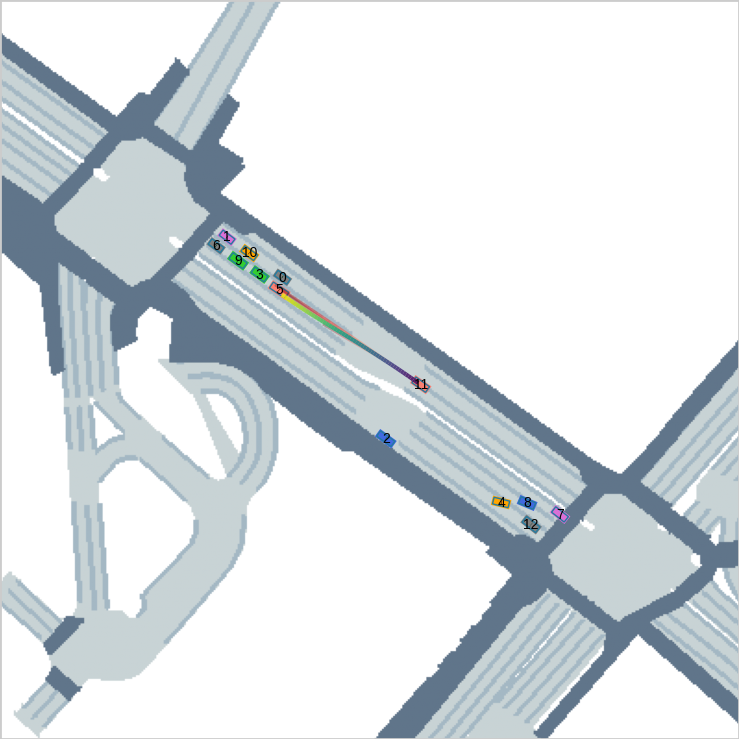}} \hfill
\caption{Qualitative results of BITS and TrafficGen fine-tuned with RM avoids unrealistic behaviors. For better visibility, we marked the unrealistic behaviors generated in {\color{red}{red} circles}.}
\label{fig:generality_example}
\end{figure}

\begin{table}[htbp]
  \centering
  \caption{Performance metrics comparison for traffic models with and without \method}
  \begin{tabular}{cccc}
    \toprule
    \textbf{Model} & \textbf{Real $\downarrow$} & \textbf{Fail $\downarrow$} & \textbf{Reward Cost$\downarrow$} \\
    \midrule
    CTG & 0.569 & 0.271 & 13.21 \\
    CTG + RM & 0.376 & 0.054	& 3.7 \\
    BITS & 1.220 & 0.314 & 12.4 \\
    BITS + RM & 0.83 & 0.23 & 9.5 \\
    TrafficGen & 1.683 & 0.39 & 15.2 \\
    TrafficGen + RM & 1.173 & 0.22 & 11.3 \\
    \bottomrule
  \end{tabular}
  \label{tab:genarality_eval}
\end{table}

\noindent\textbf{Generality of Learned RM.}
To highlight the versatility of our trained Reward Model (RM) in generalizing and enhancing a variety of models, we apply fine-tuning to two recent traffic models—BITS~\cite{bits2022} and TrafficGen~\cite{feng2022trafficgen}—using the learned RM. Given the complex designs of BITS and TrafficGen, we restrict fine-tuning to the motion forecasting components, while other model parameters remain static. For a balanced comparison, we also preserve the initialization for TrafficGen.

As depicted in Figure~\ref{fig:generality_example}, fine-tuning with RM decreases the frequency of collisions and off-road conditions in both models. We also present quantitative evaluations in Table~\ref{tab:genarality_eval}. When fine-tuned with the reward model—which was trained on traffic scenarios generated from CTG—there were measurable improvements across all three metrics for both BITS and TrafficGen. Although the degree of enhancement is not as substantial as that seen with CTG, it underscores the broad applicability of our proposed method.

\subsection{Ablation Study}
\label{sec:ablation}
In our research, we delve further into the performance capabilities of our proposed method by executing an ablation study involving the fine-tuning of the CTG model. 

\begin{table}[ht]
\centering
\caption{Ablation study on different fine-tuning strategies}
\label{tab:ablation_ft}
\begin{tabular}{l|ccc}
\toprule
                & \textbf{Real $\downarrow$} & \textbf{Fail $\downarrow$} & \textbf{Reward Cost$\downarrow$} \\
\midrule
CTG & 0.569 & 0.271 & 13.21 \\
CTG+RM (encoder)   &    0.432	&  0.176	&  9.1     \\
CTG+RM (decoder)     &    0.391	& 0.102	&	5.2     \\
CTG+RM (full)        &    \textbf{0.376}	& \textbf{ 0.054}	& \textbf{ 3.7} \\
\bottomrule
\end{tabular}
\end{table}
\noindent\textbf{Fine-tuning components.}
We experiment with fine-tuning specific components of the model and present the results in Table~\ref{tab:ablation_ft}. First, we tweak the encoder portion of the CTG model, incorporating the reward model (RM) in the process. This step leads to substantial enhancements across all performance measures, suggesting the pivotal role of human feedback, captured via the reward model, in refining the model's ability to generate realistic traffic scenarios.
For our next variant, we focus on the decoder portion of the CTG model, integrating the reward model into it. Interestingly, this setup outperforms the previous variant, reinforcing the idea that the diffusing model (decoder) might play a more integral role in the generation of more realistic traffic scenarios.
Lastly, we carry out an all-encompassing fine-tuning, involving both the encoder and decoder of the CTG model, while also adding the reward model. This configuration performs the best among all, underscoring the importance of simultaneous fine-tuning of all model components alongside the incorporation of human feedback via the reward model.
All things considered, while full fine-tuning yields the most promising results, the decoder (diffusing model) appears to be more responsive to modifications, likely owing to its inherent role in CTG's mechanism for generating realistic traffic scenarios~\cite{zhong2022guided}.

\begin{table}[h]
\centering
\caption{Ablation study on different reward models and human feedback examples.}
\label{tab:ablation}
\begin{tabular}{c|c|c|cc}
\toprule
Traffic Model               & Reward Model & Human Feedback & \textbf{Real $\downarrow$} & \textbf{Fail $\downarrow$}  \\
\midrule
\multirow{4}{*}{TrafficGen} & -            & -               & 1.683 & 0.39  \\
                            & CTG          & CTG             & 1.173 & 0.22  \\
                            & CTG          & TrafficGen      & 0.924 & 0.18  \\
                            & TrafficGen   & CTG             & \textbf{0.83}  & \textbf{0.12}  \\
                            \midrule
\multirow{4}{*}{BITS}       & -            & -               & 1.22  & 0.314 \\
                            & CTG          & CTG             & 0.83  & 0.23  \\
                            & CTG          & BITS            & \textbf{0.73}  & \textbf{0.21}  \\
                            & TrafficGen   & CTG             & 0.81  & 0.25 \\
                            \bottomrule
\end{tabular}
\end{table}

\noindent\textbf{Human feedback.} 
We experiment with training reward model using a diverse range of human feedback on traffic scenarios generated by different models. To do so, we repeated the process of collecting human feedback on generated traffic scenarios on both TrafficGen and BITS. The rest of the process, including reward model training and traffic model fine-tuning, remain the same. As the result in Table~\ref{tab:ablation}, we observe a slight improvement of realism for the fine-tuned model.

\noindent\textbf{Reward model.}
We also experiment with training reward model on different architectures. We use TrafficGen encoder as the backbone and different human feedback to train a reward model and fine-tune the other traffic models. We didn't use the BITS model due to its complicated architecture for enabling bi-level optimization. As the result shown in Table~\ref{tab:ablation}, we observe a more significant improvements on realism compared to using human feedback on different models. Based on the findings above that fine-tuning decoder being more effective, one hypothesis is that the latent representation for the traffic model and reward model should be similar and RLHF does not significantly help with learning better representations~\cite{vemprala2023chatgpt}.

%% file: body/5_conclusion.tex
\section{Conclusion}
\noindent\textbf{Summary.} In conclusion, this research has demonstrated the value of reinforcement learning with human preference (RLHF) in developing more realistic traffic models for simulation testing in autonomous vehicle development. Our method addresses the identified challenges of capturing nuanced human preferences and unifying diverse traffic simulation models. By using human feedback for alignment, we have harnessed the data efficiency of RLHF and used an autoregressive backbone model to provide a generalizable interface for the reward model input. Our TrafficRLHF model, tested on the nuScenes dataset, has proven capable of generating trajectories closely aligned with human preferences. Our contributions have not only provided the first dataset of realism alignment for traffic modeling but also offered a versatile RLHF-based framework that enhances the realism of a wide range of existing traffic models. This lays a robust foundation for future work in the field and has the potential to significantly enhance the utility of traffic simulations for autonomous vehicle development.

\noindent\textbf{Limitations and Future Work.} Currently, \method does not incorporate the more advanced fine-tuning strategies for diffusion models recently proposed~\cite{black2023training}. Furthermore, with BITS and TrafficGen, we have not yet fully harnessed the potential of \method by collecting and labeling generated traffic scenarios for these models. Training a reward model based on a specific traffic model and implementing a bespoke fine-tuning strategy could enhance traffic simulation results. Also, the iterative improvement of the traffic model through the use of updated human preference datasets and reward models might further elevate performance~\cite{vemprala2023chatgpt}. However, due to the intensive data collection required, we have not yet explored these avenues, although they present exciting opportunities for future research.

%% file: body/appendix.tex
\appendix
\section{Algorithm of Training in Details}
\noindent\textbf{CTG.}
Primarily, we adopt the training and sampling procedures outlined in \cite{zhong2022guided}, with detailed algorithms for training presented subsequently. For the reward model, we utilize an encoder from CTG and supplement it with a fully-connected neural network that comprises hidden units of varying dimensions (512-512-512-128-32) and a Tanh activation function for all layers, including the output layer. We set $\alpha=0.1$ during the fine-tuning process. Furthermore, we consistently utilize the sampling methods from \cite{zhong2022guided} for all different guidance applications.

\noindent\textbf{BITS.} Predominantly, we follow the training and sampling procedures outlined in \cite{bits2022}. For the fine-tuning process, we only adjust the spatial goal network, while the parameters for the goal-conditional policy remain unaltered. Though the goal-conditional policy network determines the actions for each agent within the traffic simulation, employing the reward model for policy network fine-tuning necessitates a different strategy, potentially resulting in an uneven comparison. Thus, we opt to fine-tune solely the spatial goal network, which also plays a critical role in guiding the traffic simulation. We set $\alpha = 0.5$ for the fine-tuning process.

\noindent\textbf{TrafficGen.} Largely, we adhere to the training and sampling procedures described in \cite{feng2022trafficgen}. Notably, TrafficGen also employs a traffic initialization network. For a fair comparison, we maintain a fixed initialization during the fine-tuning process. We set $\alpha = 0.3$ for the fine-tuning process.

All the fine-tuning process employs an amended PPO from ColossalAI~\cite{bian2021colossal} as our fine-tuning strategy with 70 epochs.

\noindent\textbf{Traffic Simulation.}
Following Zhong et al.~\cite{zhong2022guided} in CTG, to perform closed-loop traffic simulation of a scene, our model is applied for all agents in a standard control loop. At each step, the model generates a guided trajectory and executes the initial few actions before re-planning at a set frequency. In all our experiments (\Cref{sec:experiments}), each scene is rolled out for 10 seconds, starting from a ground truth driving log, and re-planned at a frequency of 2 Hz.